\documentclass[utf8]{FrontiersinVancouver} % for articles in journals using the Harvard Referencing Style (Author-Date), for Frontiers Reference Styles by Journal: https://zendesk.frontiersin.org/hc/en-us/articles/360017860337-Frontiers-Reference-Styles-by-Journal
\usepackage{url,hyperref,lineno,microtype,subcaption}
\usepackage[onehalfspacing]{setspace}
\usepackage{float} % preamble

\def\firstAuthorLast{Dinc {et~al.}} %use et al only if is more than 1 author
\def\Authors{Ugur Dinc\,$^{1,2,3*}$, Jibak Sarkar\,$^{1,2,3}$, Philipp Schubert\,$^{1,2,3}$, Sabine Semrau\,$^{1,2,3}$, Thomas Weissmann\,$^{1,2,3}$, Andre Karius\,$^{1,2,3}$, Johann Brand\,$^{1,2,3}$, Bernd-Niklas Axer\,$^{1,2,3}$, Ahmed Gomaa\,$^{1,2,3}$, Pluvio Stephan\,$^{1,2,3}$, Ishita Sheth\,$^{1,2,3}$, Sogand Beirami\,$^{1,2,3}$, Annette Schwarz\,$^{1,2,3}$, Udo Gaipl\,$^{1,2,3}$, Benjamin Frey\,$^{1,2,3}$, Christoph Bert\,$^{1,2,3}$, Stefanie Corradini\,$^{4,3}$, Rainer Fietkau\,$^{1,2,3}$ and Florian Putz\,$^{1,2,3}$}
\def\Address{$^{1}$Department of Radiation Oncology, University Hospital Erlangen, Friedrich-Alexander-Universität Erlangen-Nürnberg, Erlangen, Germany \\
$^{2}$Comprehensive Cancer Center Erlangen-EMN (CCC ER-EMN), Erlangen, Germany \\
$^{3}$Bavarian Cancer Research Center (BZKF), Munich, Germany \\
$^{4}$Department of Radiation Oncology, University Hospital, Ludwig Maximilian University of Munich, Munich, Germany}
\def\corrAuthor{Dr. Dr. Ugur Dinc}

\def\corrEmail{ugur.dinc@fau.de}

\begin{document}
\onecolumn
\firstpage{1}

\title[Benchmarking GPT-5 in Radiation Oncology]{Benchmarking GPT-5 in Radiation Oncology: Measurable Gains, but Persistent Need for Expert Oversight} 

\author[\firstAuthorLast ]{\Authors} %This field will be automatically populated
\address{} %This field will be automatically populated
\correspondance{} %This field will be automatically populated

\extraAuth{}% If there are more than 1 corresponding author, comment this line and uncomment the next one.
%\extraAuth{corresponding Author2 \\ Laboratory X2, Institute X2, Department X2, Organization X2, Street X2, City X2 , State XX2 (only USA, Canada and Australia), Zip Code2, X2 Country X2, email2@uni2.edu}

\maketitle

\begin{abstract}
\section{}
\textbf{Introduction:} Large language models (LLM) have shown great potential in clinical decision support and medical education. GPT-5 is a novel LLM system that has been specifically marketed towards oncology use. This study comprehensively benchmarks GPT-5 for the field of radiation oncology.

\textbf{Methods:} Performance was assessed using two complementary benchmarks: (i) the American College of Radiology Radiation Oncology In-Training Examination (TXIT, 2021), comprising 300 multiple-choice items, and (ii) a curated set of 60 authentic radiation oncologic vignettes representing diverse disease sites and treatment indications. For the vignette evaluation, GPT-5 was instructed to generate structured therapeutic plans and concise two-line summaries. Four board-certified radiation oncologists independently rated outputs for correctness, comprehensiveness, and hallucinations. Inter-rater reliability was quantified using Fleiss’~$\kappa$. GPT-5 results were compared to published GPT-3.5 and GPT-4 baselines.

\textbf{Results:} On the TXIT benchmark, GPT-5 achieved a mean accuracy of 92.8\%, outperforming GPT-4 (78.8\%) and GPT-3.5 (62.1\%). Domain-specific gains were most pronounced in dose specification and diagnosis. In the vignette evaluation, GPT-5’s treatment recommendations were rated highly for correctness (mean 3.24/4, 95\% CI: 3.11–3.38) and comprehensiveness (3.59/4, 95\% CI: 3.49–3.69). Hallucinations were rare (mean 10.0\%), and no case reached majority consensus for their presence. Inter-rater agreement was low (Fleiss’~$\kappa$ 0.083 for correctness), reflecting inherent variability in clinical judgment. Errors clustered in complex scenarios requiring precise trial knowledge or detailed clinical adaptation.

\textbf{Discussion:} GPT-5 clearly outperformed prior model variants on the radiation oncology multiple-choice benchmark. Although GPT-5 exhibited favorable performance in generating real-world radiation oncology treatment recommendations, correctness ratings indicate room for further improvement. While hallucinations were infrequent, the presence of substantive errors underscores that GPT-5-generated recommendations require rigorous expert oversight before clinical implementation.

Keywords: GPT-5, Artificial Intelligence, Radiation Oncology, Large Language Models, Treatment Recommendation, Oncologic Decision Support, Hallucination, Real-World Evaluation

\end{abstract}

% =========================
\section{Introduction}\label{sec:intro}

Large language models (LLMs) have advanced rapidly in recent years, driven by scaling of parameters \citep{kaplan2020scaling}, reinforcement-learning–based alignment \citep{Christiano2017_RLHF,ouyang2022training}, and the development of modular architectures such as Mixture-of-Experts (MoE) \citep{shazeer2017outrageously,fedus2022switch}. These innovations have enabled broad use of LLMs across scientific and clinical domains \citep{Vaswani2017_Attention,Brown2020_FewShot,Wei2022_CoT}. In biomedicine, domain-specific pretraining and clinical fine-tuning have enhanced representation of medical terminology and workflows \citep{Lee2020_BioBERT,Gu2021_ACMHealth,Alsentzer2019_ClinicalBERT}, while general-purpose models have achieved exam-level performance in several evaluations \citep{OpenAI2023_GPT4_Report,Singhal2023_Nature_ClinicalKnowledge,Kung2023_PLoSDigHealth_USMLE}. Nonetheless, accuracy remains heterogeneous across specialties and problem types \citep{Ebrahimi2023_IJROBP,Yalamanchili2024_JAMAOpen_LLMQuality,Chen2025_LLMsOncologyReview,Hao2025_LLMsCancerDecision}. Current consensus emphasizes transparent communication of model limitations and the need for sustained human oversight \citep{Chen2025_LLMsOncologyReview,Hao2025_LLMsCancerDecision}.

Within radiation oncology, deep learning methods are established for tasks such as segmentation, image enhancement, dose estimation, and outcome prediction \citep{huang2025principles,gomaa2025self,Huang2022_MedPhys_Mets,Weissmann2023_FrontOnc_LNLevels,Wang2023_StrontheOnkol_Cbct,Xing2020_MedPhys_DoseCalc,erdur2025deep}. LLMs extend this toolkit with text-centric applications such as guideline summarization, structured rationale generation, question answering, and automated documentation \citep{hou2025fine,Wang2023_ChatCAD,Liu2024_GPTRadPlan}. Evaluation of such models now include physics-focused question sets \citep{Holmes2023_FrontOnc_Physics,Wang2025_ShuffledPhysicsLLM} as well as surgical and board-style assessments \citep{Maruyama2025_JMIREdu_SurgicalLLM,Krumsvik2025_GPT4Assessments}. Radiation oncology-specific studies underscore both the promise of LLMs and the persistence of domain-specific limitations (e.g., dose prescription, differentiation between percutaneous and interstitial techniques) as well as challenges in keeping pace with evolving trial evidence \citep{Yalamanchili2024_JAMAOpen_LLMQuality,Chen2025_LLMsOncologyReview}.

A key area of ongoing research is the use of large language models (LLMs) as clinical decision support (CDS) systems, with prominent initiatives including Med-PaLM, Med-PaLM 2, and Google AMIE \citep{palepu2024exploring, Singhal2023_Nature_ClinicalKnowledge,singhal2025toward}. LLM agents like AMIE illustrate how LLM-based assistants may retrieve, synthesize, and contextualize medical evidence for patient-specific recommendations under expert supervision \citep{palepu2024exploring}. Early evaluations report promising accuracy in case-based reasoning and treatment planning, while underscoring the necessity of explicit uncertainty handling and clinician oversight \citep{Yalamanchili2024_JAMAOpen_LLMQuality,Hao2025_LLMsCancerDecision,Putz2025_LLMsRadOncSupport}.

GPT-5, the latest generation of OpenAI’s foundation models, represents a fundamental shift compared to GPT-3.5 and GPT-4 by explicitly incorporating reasoning-focused reinforcement learning reward models \citep{bai2022constitutional}. In combination with a larger MoE backbone and improved calibration of probabilistic outputs, GPT-5 achieves stronger logical consistency, longer-context reasoning, and higher factual accuracy. Importantly, GPT-5 is a major OpenAI model explicitly positioned as a reasoning model, designed to generate structured, interpretable rationales in addition to predictions. These advances have translated into improved performance across biomedical benchmarks, USMLE-style exams, radiology case reasoning, and oncology-specific tasks, while also reducing hallucination rates \citep{OpenAI2025_GPT5_SystemCard,OpenAI2025_GPT5_Research,Chen2025_LLMsOncologyReview,Hao2025_LLMsCancerDecision}. Despite these improvements, supervised use remains essential, particularly in high-stakes oncology settings.

Building on this progress, the present work provides the first comprehensive evaluation of GPT-5 in radiation oncology. We investigate two complementary settings:  
(i) a benchmark against the American College of Radiology Radiation Oncology In-Training Examination (ACR TXIT) subset using an automated Responses API pipeline directly comparable to GPT-3.5/4 results, and  
(ii) a novel real-world scenario dataset comprising 60 complex clinical cases without a single established standard of care \citep{Huang2023_FrontiersOncology}.  

By jointly analyzing standardized benchmarking and novel scenario-based evaluation, we assess GPT-5’s accuracy, comprehensiveness and hallucination frequency. This dual design allows rigorous quantification of performance while also examining GPT-5’s practical usability and failure modes in clinically ambiguous situations. Given the explicit positioning of GPT-5 as a reasoning model for scientific and medical tasks, our study provides a timely and domain-specific benchmark of its potential and limitations in radiation oncology.

% =========================
\section{Materials and methods}\label{sec:methods}

This study comprised two complementary evaluations of GPT-5 in radiation oncology: (i) performance on a standardized multiple-choice knowledge benchmark, and (ii) structured decision-support recommendations on real-world oncologic case vignettes. All analyses were performed on de-identified data, using isolated sessions without cross-case information transfer. No protected health information was processed. 

% -------------------------
\subsection{GPT-5 model and prompting framework}
The large language model (LLM) under test corresponds to the GPT-5 family as characterized in the publicly released system card \citep{OpenAI2025_GPT5_SystemCard}. GPT-5 is a transformer-based model trained on large text corpora with mixture-of-experts routing and reinforcement learning from human and artifical intelligence (AI) feedback. For reproducibility, standardized instructions were used for all experiments, and prompts/outputs were logged. Each API call was executed in a fresh session to avoid context leakage between cases. Automation was implemented in Python.

% -------------------------
\subsection{ACR Radiation Oncology In-Training Examination benchmark}

Knowledge-based performance was assessed using the 2021 American College of Radiology (ACR) Radiation Oncology In-Training Examination (TXIT) \citep{Rogacki2021_ARO_CarePath,Huang2023_FrontiersOncology}, which comprises 300 multiple-choice questions spanning statistics, physics, biology, and clinical radiation oncology across disease sites. Fourteen questions included medical images, of which seven required visual interpretation (Q17, Q86, Q112, Q116, Q125, Q143, Q164). Since only GPT-5 is capable of processing image-based items, these questions were included exclusively in its evaluation. For GPT-3.5 and GPT-4, the image-based items were removed prior to testing, and the total number of eligible questions was adjusted accordingly, resulting in 293 scorable items, consistent with prior work \citep{Huang2023_FrontiersOncology}.

Questions were presented as stem plus options without auxiliary text. Prompts instructed the model to select exactly one option and return the format \texttt{Final answer: X} with $X \in \{A,B,C,D\}$. No external tools (especially web search) or interactive feedback were permitted. Scoring followed established methodology: 1.0 for a correct choice and 0.0 otherwise. 

For content analysis, items were mapped to ACR knowledge domains and to a clinical care-path framework (diagnosis, treatment decision, treatment planning, prognosis, toxicity, brachytherapy, and dosimetry) \citep{Rogacki2021_ARO_CarePath}. Items explicitly referencing major clinical trials or guidelines (e.g., Stockholm~III, CRITICS, PORTEC-3, ORIOLE, AJCC 8th edition) were flagged for subgroup reporting \citep{deBoer2019_LancetOncol_PORTEC3,Phillips2020_JAMAOnc_ORIOLE,Amin2017_CA_CancerStaging8e}. Results were compared directly to published GPT-3.5 and GPT-4 benchmarks.

% -------------------------
To assess clinical usability, we curated a set of 60 anonymized oncologic case vignettes representing a broad spectrum of disease sites and treatment indications, including definitive, adjuvant, salvage, palliative, and reirradiation scenarios. Source cases were sampled from patients treated in 2025 and subsequently stripped of all identifiers. Patient name, birth date, and ID were removed automatically, while age and sex were retained. Clinical information such as diagnosis, stage, grading, comorbidities, and oncologic history was condensed by two physicians into vignettes, ensuring privacy while preserving clinical representativeness.  

The final case set was designed to be balanced across tumor sites and treatment contexts (Table~\ref{tab:case_sampling}). Specifically, it included 10 brain tumor cases (glioblastoma, lower-grade glioma, meningioma, vestibular schwannoma, paraganglioma), 10 breast cancer cases (stratified by nodal status, recurrence, and DCIS), 10 lung cancer cases (NSCLC stage~III, SBRT, reirradiation, SCLC), 10 rectal/anal cancer cases (neoadjuvant rectal, definitive anal, local recurrence), 10 prostate cancer cases (risk-adapted definitive therapy, biochemical recurrence, local recurrence post-prostatectomy), and 10 metastatic cases (brain, bone, and SBRT). This stratification allowed for evaluation of GPT-5 across both common and complex clinical scenarios, covering a representative cross-section of real-world radiation oncology practice.

\subsection{Real-world oncologic decision-support benchmark}

Each vignette was paired with a standardized instruction asking GPT-5 to propose a single most appropriate therapeutic plan and briefly justify the recommendation. Required elements included: disease stage, treatment intent, prior therapy, modality/technique, dose/fractionation, target volumes and OAR constraints, expected toxicities, and follow-up considerations. In addition, GPT-5 was instructed to generate a concise two-line summary of the proposed management, which was ultimately used for evaluation.

Outputs were evaluated by four senior radiation oncologists from a tertiary university hospital. Correctness and comprehensiveness were rated on 4-point Likert scales (4 = fully correct / comprehensive, 1 = not clinically justifiable). Hallucinations were flagged per case by each reviewer (binary). For analysis, we computed the \emph{hallucination score}, defined as the mean fraction of reviewers flagging a hallucination across cases (range 0--1). Consensus thresholds were summarized at levels of any ($\geq$1/4 raters), majority ($\geq$2/4), strong ($\geq$3/4), or unanimous (4/4). Inter-rater agreement was estimated using Fleiss’~$\kappa$ for correctness, comprehensiveness, and hallucination scores. 

Exploratory subgroup analyses stratified cases by disease site, prior radiotherapy, and treatment intent. Endpoints included correctness, comprehensiveness and hallucination rate.  

Below we provide two illustrative vignettes with GPT-5 therapy suggestions; the complete list of benchmark cases is available in the Supplementary Material. Note that dates have been removed from the published vignettes.

\textbf{Example 1 -- Case 2 (83-year-old female)} \\
\emph{Diagnosis:}
Rectal carcinoma (5--11\,cm from anal verge, 1\,cm above linea dentata), cT4b cN2b cM0, CRM+, EMVI+, G2, mismatch repair–proficient (pMMR).  

\emph{Anamnesis and work-up:}  
\#\#.\#\#.\#\#\#\#: Rectoscopy with biopsy $\rightarrow$ adenocarcinoma, G2, pMMR (MLH1, MSH2, MSH6, PMS2 intact).  
\#\#.\#\#.\#\#\#\#: CT chest/abdomen $\rightarrow$ cT3 cN+ cM0, no distant metastases.  
\#\#.\#\#.\#\#\#\#: Pelvic MRI $\rightarrow$ cT4b N2b, CRM+, EMVI+, suspected vaginal infiltration.  
\#\#.\#\#.\#\#\#\#: Multidisciplinary tumor board recommended total neoadjuvant therapy.  
\#\#.\#\#.\#\#\#\#: Transvaginal ultrasound $\rightarrow$ no sonographic vaginal invasion, fat plane preserved.  

\emph{GPT-5 suggestion (TNT, long course):}  
Radiochemotherapy with 50.4\,Gy/28 fractions (IMRT) plus capecitabine 825\,mg/m\textsuperscript{2} BID (Mon--Fri),  
consolidation chemotherapy with mFOLFOX6 q14d $\times$4 cycles,  
followed by total mesorectal excision (TME) 8--12 weeks post-RCT.

\textbf{Example 2 -- Case 52 (male, 52 years).}  \\
\emph{Diagnosis:}
Prostate carcinoma cT1c cN0 cM0, Gleason 4+3=7b, iPSA 10.8\,ng/ml, estimated lymph node risk (YALE) $\sim$7\%.  

\emph{Anamnesis and work-up:}  
\#\#.\#\#.\#\#\#\#: Prostate biopsies $\rightarrow$ poorly differentiated adenocarcinoma, max.\ Gleason 7b (4+3).  
Since \#\#.\#\#\#\#: LHRH 3-monthly depot.  
\#\#.\#\#.\#\#\#\#: Tumor board (urologic cancers) $\rightarrow$ recommendation for local curative therapy, patient in good general condition.  
\#\#.\#\#.\#\#\#\#: Prostate MRI (pelvis) $\rightarrow$ no suspicious lymph nodes, indeterminate bone lesion (likely atypical hemangioma), MRI performed for radiotherapy planning.  
\#\#.\#\#.\#\#\#\#: Uncomplicated implantation of 3 fiducial gold markers.  

\emph{GPT-5 suggestion (curative RT + short-term ADT):}  
Definitive radiotherapy to prostate $\pm$ proximal seminal vesicles, 60\,Gy in 20 fractions (IG-IMRT/VMAT with daily IGRT);  
no elective pelvic lymph node irradiation;  
short-term ADT for 4–6 months (if already $>$6 months, terminate now).

% -------------------------
\subsection{Statistical analysis}
For the TXIT benchmark, we report overall accuracy and descriptive breakdowns by domain, care-path category, and trial/guideline-anchored items. For the clinical decision-support benchmark, primary endpoints were expert-rated correctness, comprehensiveness, hallucinations, and concordance with delivered care. Exploratory subgroup analyses were prespecified; no multiplicity adjustment was applied. All automation, randomization seeds, prompts, and raw outputs are provided as Supplementary Material. Use of de-identified, retrospective vignettes complied with institutional policies for research on non-human-subjects data.

% =========================
\section{Results}\label{sec:results}

\subsection{Overall TXIT performance}
Across the 293 scorable items of the ACR TXIT (2021), previously reported baselines reproduced web–interface performance of 63.1\% for GPT-3.5 and 74.1\% for GPT-4. Using the application programming interface (API) with a fixed prompt over five repeated runs, GPT-3.5 achieved $62.1\%\pm1.1\%$ and GPT-4 $78.8\%\pm0.9\%$, consistent with earlier reports \citep{Huang2023_FrontiersOncology}. 

For GPT-5, we conducted five independent runs that included both text-only and image-based questions. Overall accuracy ranged from 92.3\% to 93.0\%, with a mean of 92.8\%. Performance on the subset of image-based items was lower, with only 2 of 7 questions answered correctly. Given that the item pool, scoring criteria, and adjudication procedures were identical to those used for prior models, the observed improvement in accuracy reflects genuine advances in model capability rather than differences in test format or evaluation methodology.

\subsection{Domain-wise analysis}
Stratified by ACR knowledge domains, GPT-5 preserved historical strengths, reaching at least 95\% accuracy in Statistics, CNS/Eye, Biology, and Physics. Performance remained lower for Gynecology (75.0\%), and moderately reduced for Gastrointestinal and Genitourinary topics (both around 90\%). Compared with GPT-4, the largest absolute gains were observed in Dose (from 59.4\% to 87.5\%) and Diagnosis (from 76.5\% to 91.2\%).

\subsection{Clinical care-path analysis}
When mapped to clinical care-path categories, GPT-5 demonstrated consistently high accuracy:
\begin{itemize}
    \item 100\% in Treatment Planning (11/11), Local Control (2/2), Diagnosis Methodology (3/3), Anatomy (8/8), and Pharmacology (1/1).
    \item 95.9\% (47/49) in Treatment Decision and 95.2\% (20/21) in Prognosis Assessment.
    \item 92.3\% in Toxicity, 92.9\% in Trial/Study/Guideline, and 91.2\% in Diagnosis.
    \item 88.9\% in Brachytherapy and 87.5\% in Dose.
\end{itemize}

On items explicitly anchored to named trials or staging systems, GPT-5 achieved 92.9\% (13/14), outperforming GPT-4 (85.7\%) and GPT-3.5 (50.0\%).

\subsection{Evaluation on clinical case vignettes}
In the evaluation of 60 real-world radiation oncology vignettes, GPT-5’s treatment recommendations were rated as follows:
\begin{itemize}
    \item Correctness: mean 3.24/4 (95\% CI: 3.11–3.38).
    \item Comprehensiveness: mean 3.59/4 (95\% CI: 3.49–3.69).
\end{itemize}
Hallucinations were infrequent. In total, 24 of 240 individual ratings (60 cases × 4 raters) were classified as hallucinations, corresponding to an overall hallucination rate of 10\%. Thus, the vast majority of ratings (90\%) did not identify hallucinations, indicating that such occurrences were infrequent across cases. No case was flagged by the majority of experts (at least two out of four raters). Distribution was 36/60 (60\%) with zero hallucination flags and 24/60 (40\%) with exactly one.

Inter-rater reliability was low, with Fleiss’ $\kappa=0.083$ for correctness, $\kappa=-0.016$ for comprehensiveness, and $\kappa=-0.111$ for hallucinations, indicating variability in individual reviewer judgments.

\begin{figure}[H]
\centering
\includegraphics[width=0.85\linewidth]{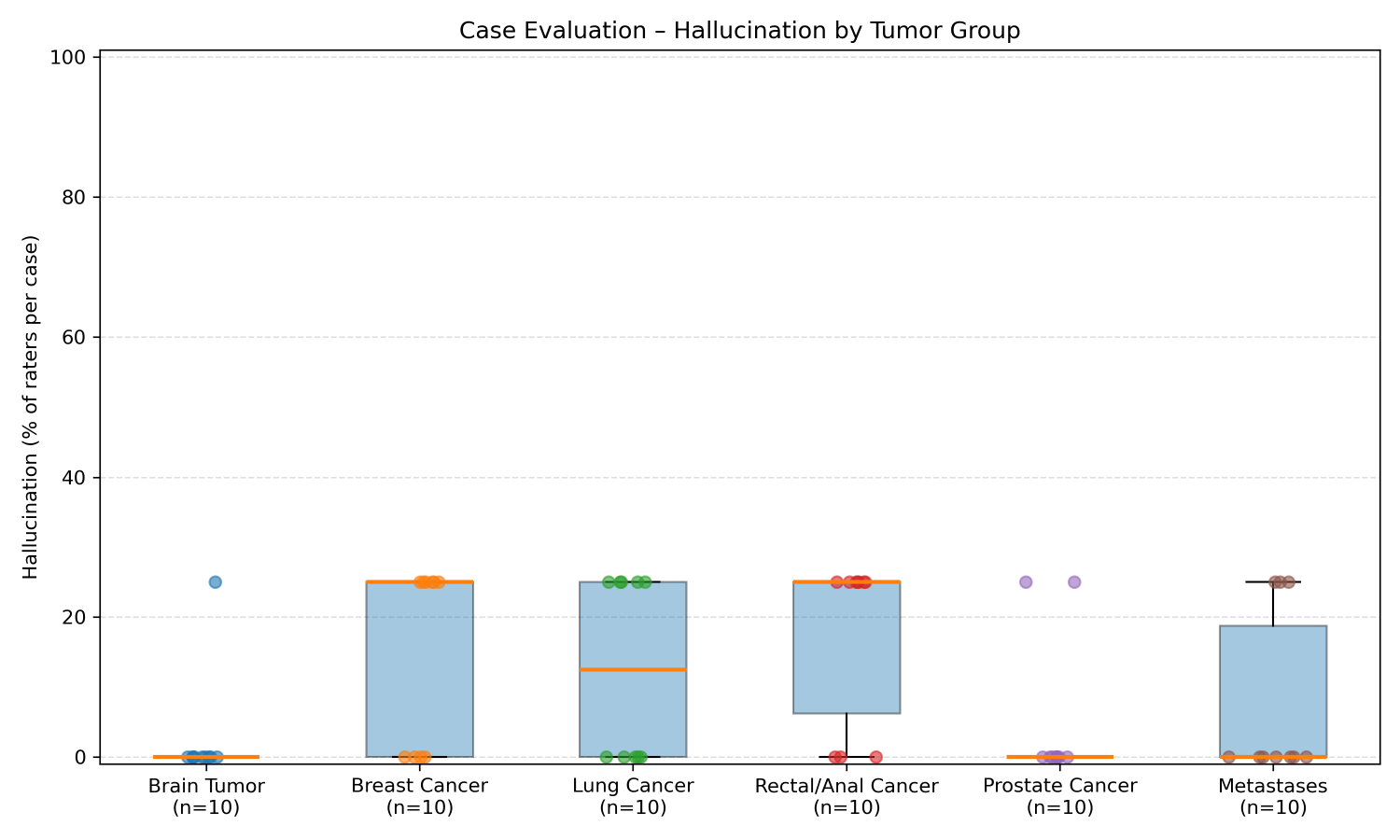}
\caption{Hallucination rates by tumor group (10 cases each).  
Hallucinations were rare overall. Prostate and brain tumor cases showed almost no hallucinations, while breast, rectal/anal, lung, and metastasis cases exhibited higher variability, with some reaching up to 25\%.}
\label{fig:hallucination_box}
\end{figure}

\begin{figure}[H]
\centering
\includegraphics[width=0.85\linewidth]{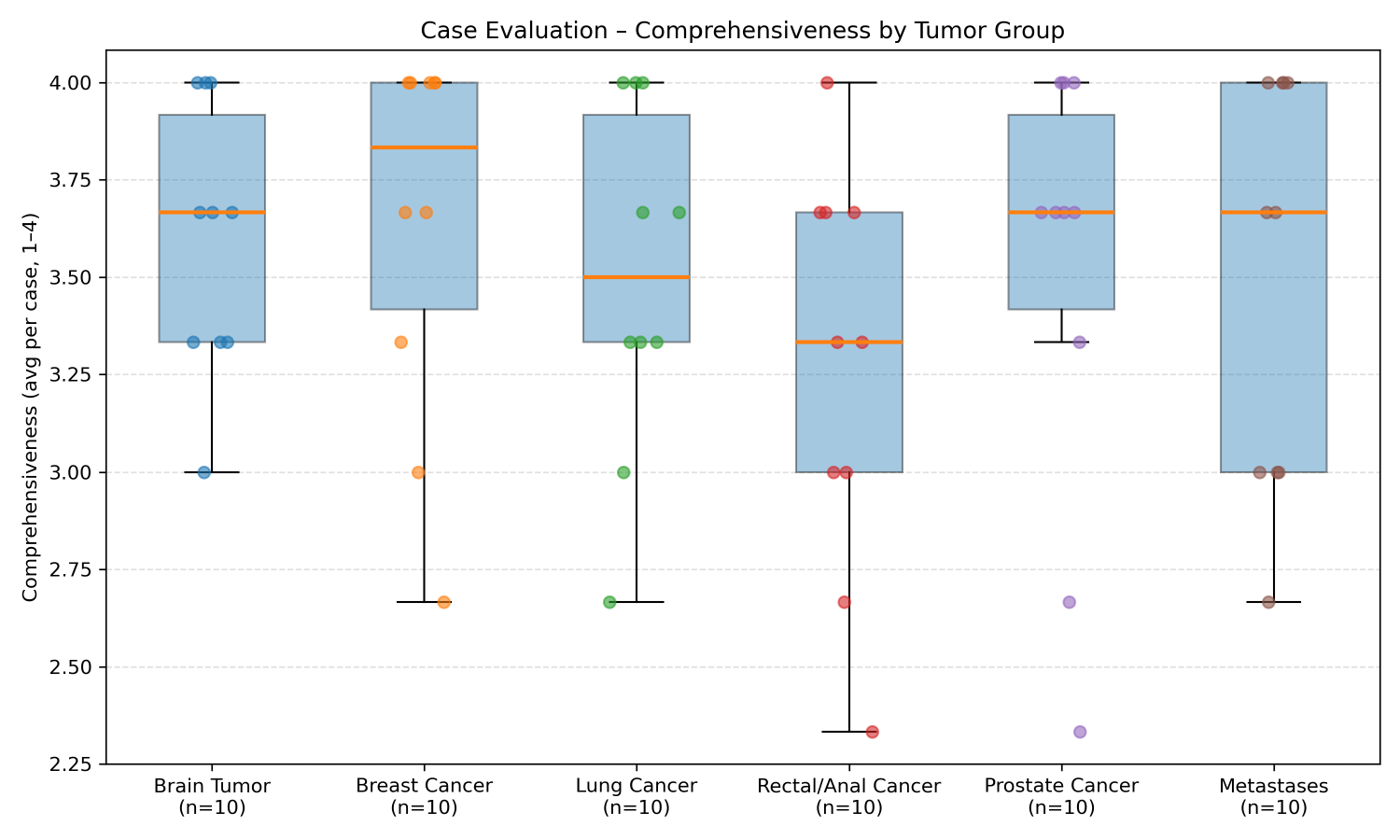}
\caption{Comprehensiveness ratings (1--4 scale) by tumor group.  
All groups except for rectal/anal achieved high median scores ($\geq$3.5). Breast, prostate, and brain tumor cases were most consistently rated as highly comprehensive, while rectal/anal and lung cancers showed broader variability.}
\label{fig:comprehensiveness_box}
\end{figure}

\begin{figure}[H]
\centering
\includegraphics[width=0.85\linewidth]{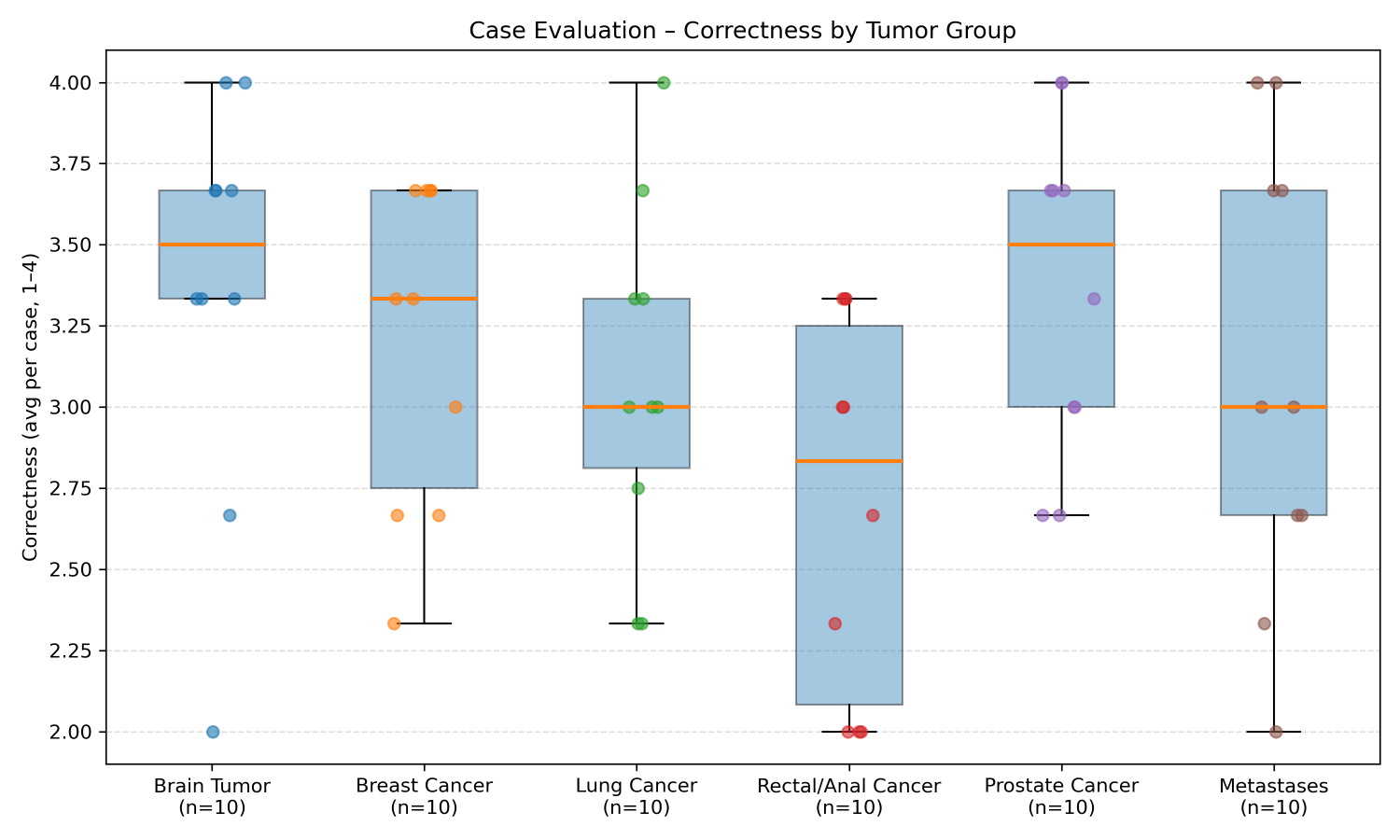}
\caption{Correctness ratings (1--4 scale) by tumor group.  
Prostate and brain tumor cases scored highest for correctness (median $\geq$3.5). Lung and rectal/anal cancer cases showed lower and more variable correctness, whereas breast cancer and metastasis cases performed intermediately.}
\label{fig:correctness_box}
\end{figure}

When stratified by the six major tumor groups, distinct patterns emerged (Figures~\ref{fig:hallucination_box}--\ref{fig:correctness_box}).  
Hallucinations were rare overall, with prostate and brain tumor cases showing almost none, whereas breast, rectal/anal, lung, and metastasis cases exhibited higher variability (Figure~\ref{fig:hallucination_box}).  
Comprehensiveness was generally high across all groups (median $\geq$3.5/4), with breast, prostate, and brain tumors rated most consistently complete, and rectal/anal and lung cancers showing broader variability (Figure~\ref{fig:comprehensiveness_box}).  
Correctness displayed the clearest differentiation: prostate and brain tumors achieved the highest median scores ($\geq$3.5/4), breast and metastases performed intermediately, while rectal/anal and lung cancers scored lowest and most variably (Figure~\ref{fig:correctness_box}).

Beyond these main groupings, more granular subgroup analyses revealed clear data-defined differences. Highest correctness was observed in \emph{prostate, intermediate risk} (correctness 3.83; comprehensiveness 3.83; hallucinations 0\%) and \emph{prostate, biochemical recurrence after RPE} (3.67; 3.83; 0\%), as well as \emph{small-cell lung cancer (SCLC)} (3.67; 4.00; 0\%). Brain primaries were generally strong: \emph{meningioma} (3.67; 3.67; 0\%), \emph{vestibular schwannoma} (3.67; 3.33; 0\%), \emph{glioma grade 2/3} (3.50; 3.67; 0\%), and \emph{glioblastoma} (3.33; 3.83; 0\%); a weaker brain subgroup was \emph{pituitary adenoma} (2.67; 3.50; 12.5\%). 

Breast adjuvant scenarios were solid for correctness and highly complete but showed higher hallucination rates in several subgroups: \emph{adjuvant low risk} (3.33; 4.00; 25\%), \emph{adjuvant node-negative} (3.50; 4.00; 25\%), \emph{adjuvant node-positive} (3.17; 3.50; 12.5\%), \emph{DCIS} (3.00; 3.33; 12.5\%), and \emph{loco-regional recurrence} (3.00; 3.33; 0\%). 

Within lung cancer, \emph{NSCLC stage III (definitive)} performed moderately (3.33; 3.44; 8.33\%), whereas settings requiring finer adaptation were lower: \emph{NSCLC re-irradiation} (2.54; 3.50; 12.5\%) and \emph{NSCLC SBRT} (2.78; 3.22; 25\%). 

For metastatic disease, \emph{palliative bone metastases} performed well (3.67; 3.89; 0\%) and \emph{metastatic SBRT} was acceptable (3.11; 3.33; 0\%), while \emph{brain metastases} were lower and more error-prone (2.67; 3.33; 18.75\%). 

Rectum–anal cases were the weakest overall: \emph{anal cancer} (3.00; 3.00; 0\%), \emph{neoadjuvant rectal cancer} (2.75; 3.25; 25\%), and \emph{local recurrence} (2.33; 3.56; 25\%).

These patterns localize remaining challenges to problem settings that demand precise trial knowledge, dose/fractionation choices, or complex multimodality sequencing.

Several representative cases illustrate these limitations:

\begin{itemize}
    \item Case 7 (low-risk prostate cancer): The system recommended definitive therapy, which some raters considered overtreatment given active surveillance would also have been guideline-concordant, though others accepted it because the patient requested therapy.  
    \item Case 8 (neoadjuvant rectal cancer): Biomarker analysis (e.g. MSI) was omitted, a gap increasingly relevant for therapy planning.  
    \item Case 11 (brain metastases): The combination of ipilimumab and nivolumab was proposed, but with a non–guideline-concordant dosing scheme, lowering correctness.  
    \item Case 17 (DCIS): A radiotherapy boost dose was recommended, which is not guideline-supported and was judged overtreatment.  
    \item Case 29 (NSCLC with SBRT): Systemic therapy options were suggested despite definitive SBRT being the standard in this context.  
    \item Case 36 (lung SBRT): Chemotherapy cycles were not specified, reducing precision despite an otherwise acceptable plan.  
    \item Case 42 (brain metastases): Therapy recommendations were given without histologic confirmation, a prerequisite step that reduced guideline adherence.  
    \item Case 58 (SBRT): Outdated regimens were mixed with correct ones, producing polarized ratings.  
\end{itemize}

Where historical records permitted, GPT-5’s recommendations showed high concordance with delivered care across treatment intent, modality/technique, and dose/target ranges, although multiple reasonable options often existed.

\section{Discussion}\label{sec:discussion}

This work extends the literature on large language models (LLMs) in radiation oncology along two complementary axes: standardized examination performance and real-world decision support. On the ACR TXIT subset, the present model attained 92.8\% under the same item pool and adjudication rules previously used for GPT\mbox{-}3.5/4, which yielded  63.1\% and  74.1\%  respectively \citep{Huang2023_FrontiersOncology}. The magnitude of this gain is consistent with the architectural and training changes that distinguish GPT-5 from its predecessors: scaling of transformer capacity, more stable long-context attention, preference optimization with richer feedback, and—critically—its explicit positioning as a reasoning model. Unlike GPT-3.5 and GPT-4, GPT-5 is designed not only to recall information but to generate structured, stepwise rationales, yielding more consistent and interpretable outputs \citep{OpenAI2025_GPT5_SystemCard,OpenAI2025_GPT5_Research}. Our use of a constrained response format (``Final answer: X'') reduced adjudication noise without contributing domain knowledge. Examination accuracy nevertheless remains an imperfect surrogate for clinical competence: persistent weaknesses were observed in brachytherapy, fine-grained dosimetry, and evolving trial-specific details, mirroring deficits documented for earlier models \citep{Huang2023_FrontiersOncology}.

Comparisons with adjacent evaluations clarify where improvements reflect reasoning advances rather than test artifacts. In radiation oncology physics, Holmes \emph{et~al.} showed that item structure and distractors meaningfully influence performance \citep{Holmes2023_FrontOnc_Physics}, while Wang \emph{et~al.} demonstrated that simply shuffling answer options alters accuracy \citep{Wang2025_ShuffledPhysicsLLM}. Our reproduction of prior TXIT baselines with identical stems, options, and scoring therefore supports the interpretation that GPT-5’s higher scores reflect genuine advances in reasoning and knowledge synthesis rather than format effects. At the same time, gynecologic oncology, brachytherapy, and trial-anchored items remained more challenging, consistent with topic-level variability in prior reports \citep{Huang2023_FrontiersOncology,Yalamanchili2024_JAMAOpen_LLMQuality,Ebrahimi2023_IJROBP}.

A more robust and clinically relevant benchmark was established through our 60-case evaluation based on authentic oncologic vignettes, in which GPT-5 was tasked with generating structured management plans. The model produced coherent and comprehensive drafts, extending earlier findings with GPT-4 that had been reported in a smaller Red Journal–style Gray Zone cases \citep{Huang2023_FrontiersOncology}. Hallucinations were infrequent and did not pose a substantive concern; rather, errors were concentrated in areas requiring detailed trial-specific knowledge,  nuanced clinical adaptation or complex multimodality treatments (e.g., SBRT, DCIS, brain metastases, ano-rectal cancer, lung cancer with comorbidities). These results highlight the potential of reasoning-oriented models: GPT-5 is able to synthesize case-relevant rationales, thereby moving closer to providing the deliberative support that is directly applicable in tumor board settings.

Our evaluation complements emerging work such as the \emph{Articulate Medical Intelligence Explorer (AMIE)} system, which was tested in synthetic breast oncology vignettes \citep{palepu2024exploring}. While AMIE incorporated retrieval and self-critique pipelines and demonstrated performance above trainees and fellows, our study extends this line of research by benchmarking GPT-5 on real, anonymized, multi-disease radiation oncology cases rated by board-certified specialists.

Our findings align with broader medical LLM studies, which consistently show topic-level heterogeneity, stronger outputs under expert scaffolding, and improved interpretability when reasoning steps are made explicit \citep{Yalamanchili2024_JAMAOpen_LLMQuality,Ebrahimi2023_IJROBP,Krumsvik2025_GPT4Assessments,Maruyama2025_JMIREdu_SurgicalLLM}. Reviews and meta-analyses converge on supervised applications—education, tumor-board summarization, pre-board preparation—rather than autonomous decision-making \citep{Hao2025_LLMsCancerDecision,Chen2025_LLMsOncologyReview, trapp2025patient}. Within this trajectory, GPT-5’s positioning as a reasoning model represents a qualitative step: it enables explicit rationale generation and structured synthesis across complex oncology cases, something prior models handled only inconsistently.

Methodologically, our study also connects to emerging multimodal planning assistants that couple LLM reasoning to imaging or dose engines \citep{Wang2023_ChatCAD,Liu2024_GPTRadPlan}. Such systems may compensate for GPT-5’s current blind spots in image review and dosimetry but will require rigorous validation and regulatory oversight before clinical use \citep{Ebrahimi2023_IJROBP,Gilbert2023_NatMed_Devices}. From a governance perspective, near-term deployment should remain supervised, with retrieval-augmented pipelines, auditable links to guidelines and trials (e.g., AJCC, PORTEC-3, ORIOLE), explicit uncertainty labeling, and human sign-off \citep{Liu2023_JAMIA_CDS,Amin2017_CA_CancerStaging8e,deBoer2019_LancetOncol_PORTEC3,Phillips2020_JAMAOnc_ORIOLE,Gilbert2023_NatMed_Devices}.

Several limitations frame the interpretation of our findings. First, standardized examinations sample knowledge in a manner that differs from real-world practice; thus, high TXIT performance does not guarantee reliability in rare, ambiguous, or evolving scenarios \citep{Singhal2023_Nature_ClinicalKnowledge}. Second, the retrospective case cohort—while enriched for evaluability through follow-up and therapy verification—may be affected by survivorship and documentation biases. Third, despite the use of prespecified rubrics, inter-rater variability persisted, typical for complex oncologic vignettes \citep{Chen2025_LLMsOncologyReview}. Fourth, our evaluation did not incorporate tool use or external retrieval, which could plausibly further improve accuracy through real-time access to guidelines and trial data. Fifth, the TXIT and 60-case sets cover only a subset of radiation oncology knowledge, and therefore cannot capture the full diversity of clinical decision-making. Finally, model versioning, decoding parameters, and chat memory constraints influence outputs; we mitigated these factors through fresh conversations and repeated runs, but residual variability remains. 

Future research should prioritize prospective studies, ideally randomizing tumor-board workflows to model-assisted versus control arms. Additional directions include systematic evaluation of reasoning under tool use and retrieval conditions, direct comparisons of general-purpose and domain-adapted models, explicit assessment of target and dose coherence as well as toxicity forecasting, and development of auditable retrieval pipelines tightly coupled to primary evidence \citep{Hao2025_LLMsCancerDecision}. Ultimately, the goal is not to replace clinician judgment, but to generate reproducible, evidence-linked drafts and option sets that accelerate multidisciplinary deliberation while preserving safety, accountability, and trust.

\section{Limitations}\label{sec:limitations}
Limitations include the absence of tool use and external retrieval, which could further improve performance by enabling real-time access to guidelines, trial updates, and dose–constraint references. The TXIT and 60-case sets cover only a subset of radiation oncology knowledge and therefore cannot capture the full spectrum of clinical decision-making. Outcomes are also influenced by model versioning, decoding parameters, and chat memory constraints; we mitigated these effects through fresh conversations and repeated runs (five each) for GPT-3.5, GPT-4, and GPT-5, but residual variability remains.

\section{Conclusion}\label{sec:conclusion}

On TXIT benchmarks, GPT-4 outperformed GPT-3.5, and \textbf{GPT-5 further increased accuracy to 92.8\%}, with strengths in statistics, CNS/eye, physics, diagnostic methods, and toxicity, and persistent gaps in gynecology, brachytherapy, dosimetry, and trial-specific details. 

More importantly, our novel 60-case benchmark of real-world oncologic scenarios showed that GPT-5 can generate coherent, comprehensive management drafts. Hallucinations were rare and not a substantive concern; limitations instead reflected occasional inaccuracies in guideline details, multi-modal treatments, and nuanced adaptation of medical knowledge to complex clinical case descriptions. 

Overall, GPT-5 emerges as a reasoning model with value in supervised applications such as education, pre-board preparation, and draft generation for tumor boards. Its near-term role is as an augmentative assistant, with human verification and evidence retrieval remaining essential safeguards.

\section*{Conflict of Interest Statement}
The authors declare no commercial or financial relationships that could be construed as a potential conflict of interest.

\section*{Data Availability Statement}
All prompts, grading rules, and aggregated results (per-run accuracies) are shared as Supplementary Material. The ACR TXIT 2021 exam is publicly accessible from ACR (usage per their terms).

\section*{Ethics statement}
The use of de-identified, case vignettes complied with institutional research policies and local legislation (BayKrG Art. 27) as well as with the Helsinki declaration and its subsequent amendments. All patients had given their full consent to the use of patient data for secondary scientific purposes. Patient case vignettes had been stripped of all identifiers. Moreover, clinical information such as diagnosis, stage, grading, comorbidities, and oncologic history was condensed by two physicians into vignettes, enabling full privacy while preserving clinical representativeness.

\bibliographystyle{Frontiers-Vancouver} %  Many Frontiers journals use the Harvard referencing system (Author-date), to find the style and resources for the journal you are submitting to: https://zendesk.frontiersin.org/hc/en-us/articles/360017860337-Frontiers-Reference-Styles-by-Journal. For Humanities and Social Sciences articles please include page numbers in the in-text citations 
\bibliography{test}

%%% Make sure to upload the bib file along with the tex file and PDF
%%% Please see the test.bib file for some examples of references
\bigskip
\bigskip
\bigskip
\bigskip
\section*{Figure captions}

%%% Please be aware that for original research articles we only permit a combined number of 15 figures and tables, one figure with multiple subfigures will count as only one figure.
%%% Use this if adding the figures directly in the mansucript, if so, please remember to also upload the files when submitting your article

\begin{figure}[H]
\centering
\includegraphics[width=\linewidth]{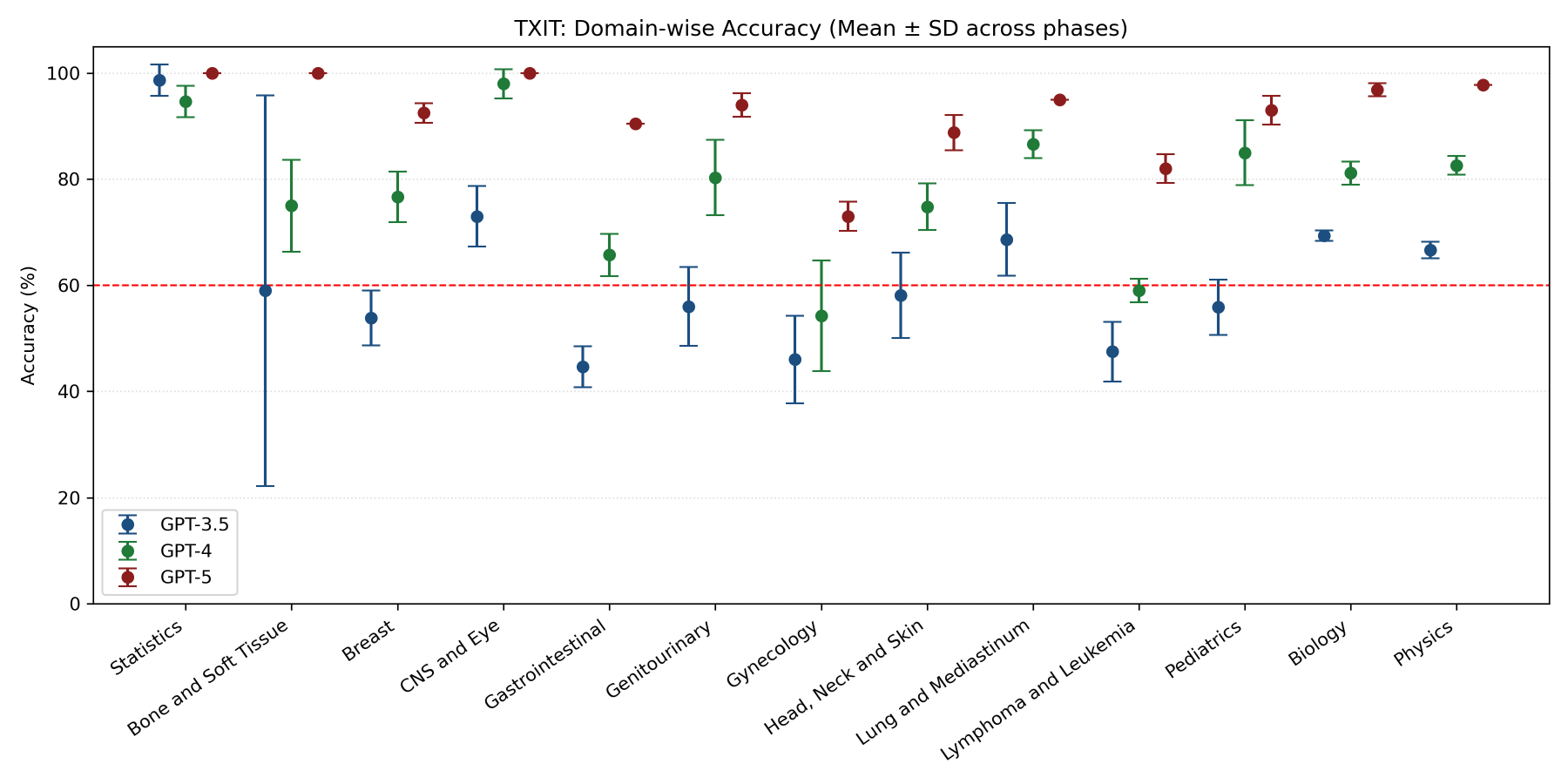} 
\caption{TXIT accuracy by model. Symbols show mean accuracy and error bars indicate the standard deviation (SD) across five runs for GPT-3.5, GPT-4, and GPT-5.}
\label{fig:overall}
\end{figure}

\newpage

\begin{figure}[H]
\centering
\includegraphics[width=\linewidth]{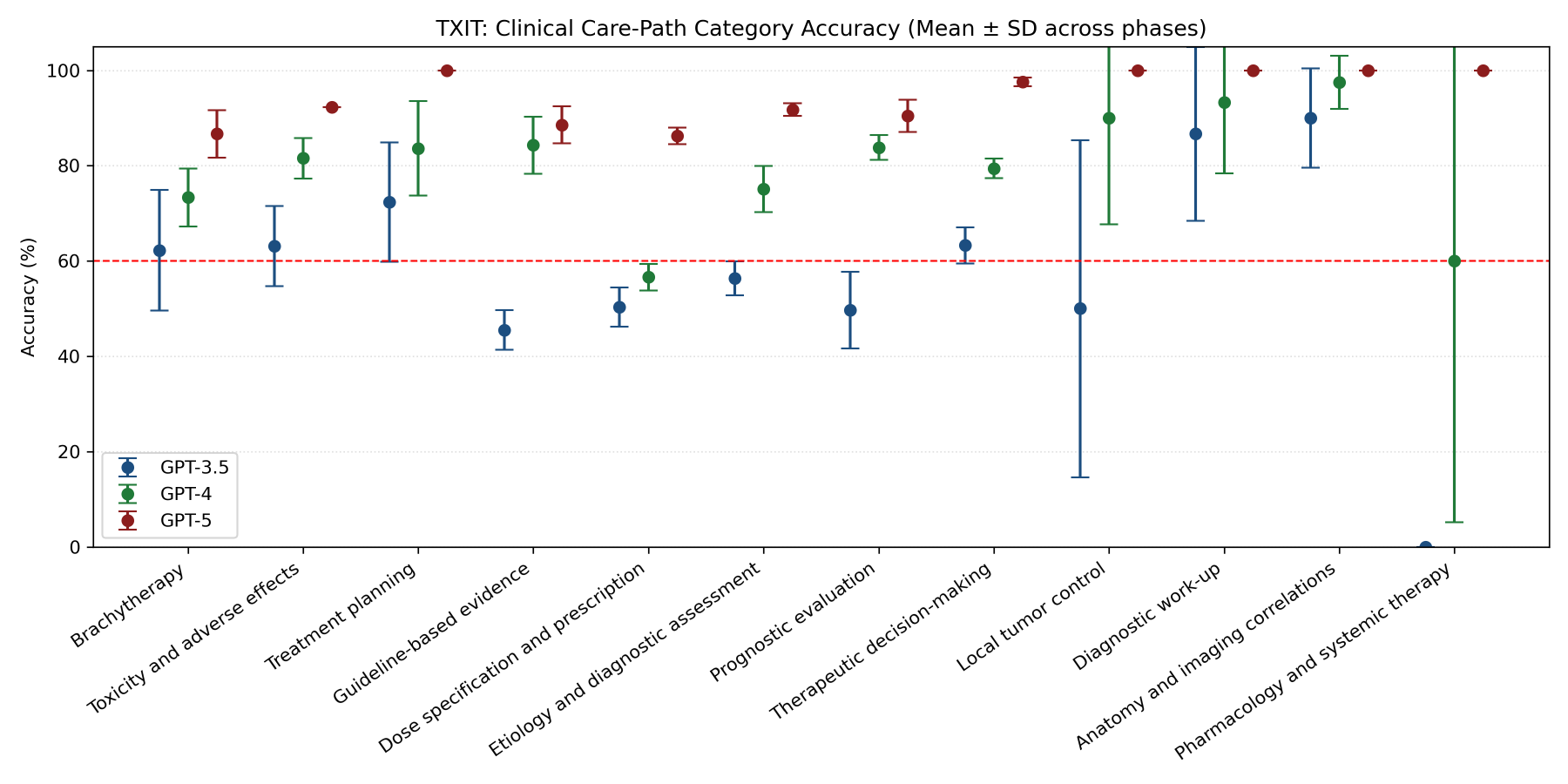} 
\caption{Domain-wise accuracy across models. Symbols show mean accuracy and error bars indicate the SD across five runs for GPT-3.5, GPT-4, and GPT-5.}
\label{fig:domains}
\end{figure}

\begin{figure}[H]
\centering
\includegraphics[width=0.72\linewidth]{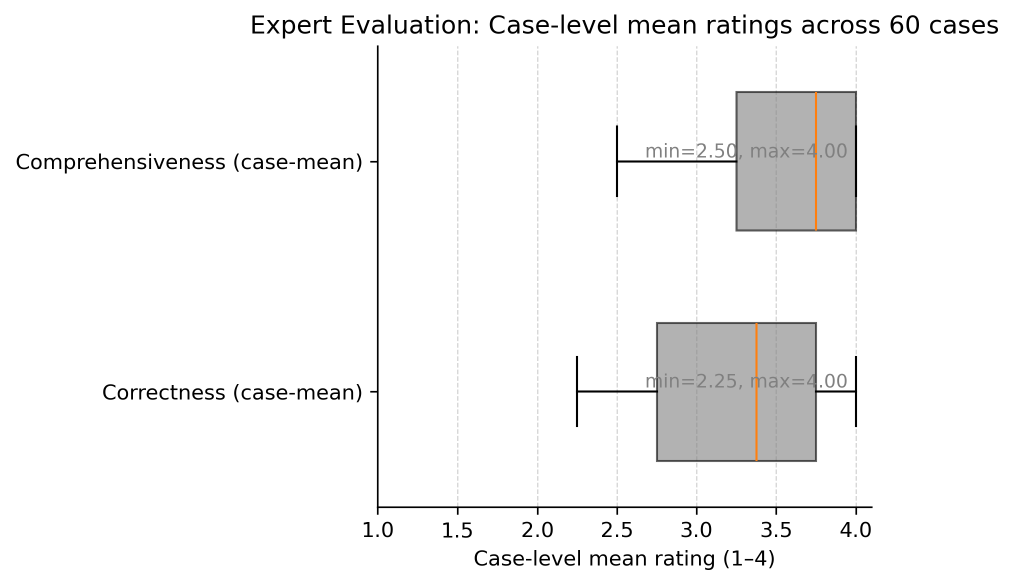}
\caption{Distribution of case-level mean expert ratings for correctness and comprehensiveness across 60 cases. Each box represents the inter-quartile range (IQR) with whiskers indicating outliers, summarizing the distribution of ratings. Case-level mean correctness ranged from 2.25 to 4.00, while case-level mean comprehensiveness ranged from 2.50 to 4.00.}
\label{fig:gz_points}
\end{figure}
\newpage
\begin{figure}[H]
\centering
\includegraphics[width=0.8\linewidth]{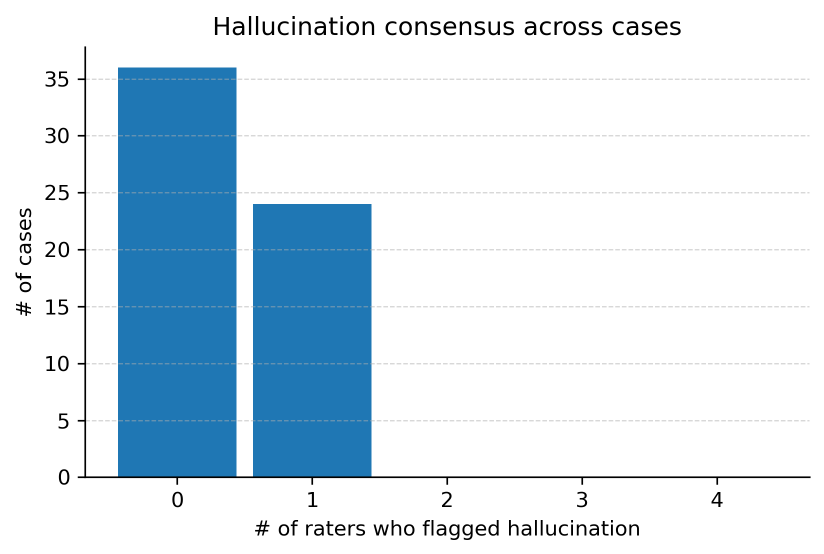}
\caption{Hallucination consensus across cases. Bars show the number of cases with 0, 1, 2, 3, or 4 raters flagging hallucination. In this cohort, 36/60 cases had 0 flags and 24/60 had exactly 1 flag; no case reached majority ($\geq$2/4), strong ($\geq$3/4), or unanimous (4/4) consensus.}
\label{fig:hall_hist}
\end{figure}

\begin{table}[H]
\centering
\caption{Sampling frame for the 60 clinical cases set. Category totals were balanced to support stratified analyses by site and radiooncologic treatment indication.}
\label{tab:case_sampling}
\begin{tabular}{llc}
\hline
\textbf{Category} & \textbf{Subcategory} & \textbf{Number of cases} \\
\hline
Brain Tumor & Glioblastoma (grade 4) & 2 \\
            & Glioma (grade 2/3)     & 2 \\
            & Meningioma              & 2 \\
            & Vestibular schwannoma   & 2 \\
            & Paraganglioma           & 2 \\
\hline
Breast Cancer & Adjuvant—nodal positive & 2 \\
              & Adjuvant—nodal negative & 2 \\
              & Adjuvant—low risk        & 2 \\
              & Loco-regional recurrence & 2 \\
              & DCIS                     & 2 \\
\hline
Lung Cancer  & NSCLC—definitive (stage III) & 3 \\
             & NSCLC—SBRT                   & 3 \\
             & NSCLC—reirradiation          & 2 \\
             & SCLC                         & 2 \\
\hline
Rectal/Anal Cancer & Neoadjuvant—rectal cancer & 4 \\
                   & Definitive—anal cancer      & 3 \\
                   & Local recurrence             & 3 \\
\hline
Prostate Cancer & Definitive—low risk              & 2 \\
                & Definitive—intermediate risk     & 2 \\
                & Definitive—high risk             & 2 \\
                & Biochemical recurrence after RPE  & 2 \\
                & Local recurrence after RPE + EBRT & 2 \\
\hline
Metastases     & Brain metastases           & 4 \\
               & Palliative—bone metastases & 3 \\
               & SBRT                       & 3 \\
\hline
\end{tabular}

\vspace{0.5em}
\footnotesize \textbf{Abbreviations:} DCIS, ductal carcinoma in situ; NSCLC, non–small cell lung cancer; SCLC, small-cell lung cancer; SBRT, stereotactic body radiotherapy; RPE, radical prostatectomy; EBRT, external beam radiotherapy.
\end{table}

\begin{table}[H]
\centering
\caption{Clinical cases : case-level outcomes across raters.}
\label{tab:gz}
\begin{tabular}{lccc}
\hline
Metric & Estimate & 95\% CI & Notes \\
\hline
Correctness (mean/4) & \textbf{3.24} & \textbf{3.11–3.38} & mean across cases \\
Comprehensiveness (mean/4) & \textbf{3.59} & \textbf{3.49–3.69} & mean across cases \\
Hallucination (mean \% per case) & \textbf{10.0\%} & \textbf{6.8–13.2\%} & proportion of raters per case \\
Any hallucination (per case) & \textbf{40.0\%} & \textbf{28.6–52.6\%} & 24/60 cases \\
Majority/Strong/Unanimous & \textbf{0\%/0\%/0\%} & -- & no case $\geq$2/4, $\geq$3/4, or 4/4 \\
Inter-rater reliability (Fleiss’ $\kappa$) & 0.083 / -0.016 / -0.111 & -- & correctness / compreh. / hallucination \\
\hline
\end{tabular}
\end{table}

%%% If you don't add the figures in the LaTeX files, please upload them when submitting the article.
%%% Frontiers will add the figures at the end of the provisional pdf automatically
%%% The use of LaTeX coding to draw Diagrams/Figures/Structures should be avoided. They should be external callouts including graphics.
\newpage
\section*{Supplementary Material}

\begin{verbatim}

 f"""
You are a tumor board assistant in Germany (radiation oncology, medical oncology, surgical, ENT).
Cite German S3 guidelines first with exact recommendation numbers and short direct quotes.
Use secondary sources (NCCN/ESMO/ESTRO/ICRU) only as supportive.

Output ONE SINGLE LINE of JSON with EXACTLY these keys:

- "diagnosis_compact": three short lines, each very concise:
    Line 1: cancer type + side/site (if known) + ED:MM/YY (Erstdiagnose), and if applicable: Rez MM/YY (Rezidivmonat/-jahr)
        e.g., "Mammakarzinom links ED:05/2023, Rez 03/2025"
    Line 2: TNM (c/p + T,N,M), key biology (e.g., ER/PR, HER2, p16/HPV, RAS/BRAF, MSI/MSS, Gx)
        e.g., "cT2 cN1 cM0, ER/PR+, HER2–, G2"
    Line 3: starts with "Bisherige onkologische Therapie:" + last relevant procedure/systemic therapy, very compact
        e.g., "Bisherige onkologische Therapie: Zn. OP (BET) 06/2025" or "Bisherige onkologische Therapie: Zn. 6x Cis/Pembro"
- "therapy_compact": one concise line (German abbreviations) like:
    "adj. RCT: 50 Gy/25 Fr (ED 2.0 Gy) + Boost 10 Gy/5 Fr; Chemo: CAPOX q21d ×3"
- "tumorformel": concise TNM/tumor formula if inferable; otherwise "Unklar".
- "suggested_therapieplan": concrete, guideline-aligned plan. If listing multiple points,
  put each point on its own line starting with a number and a closing parenthesis,
  e.g., "1) First thing\\n2) Second thing\\n3) Third thing".
  Cover: chemotherapy (drug(s), schema, dose or range), radiotherapy (technique, target volumes, total dose & fractionation),
  surgery (if/when appropriate), tumorboard recommendation.
- "notes_to_clinician": practical next steps (further diagnostics before/after/during, labs, imaging, pathology, p16/HPV if HNSCC,
  renal/hepatic clearance, dental eval, PEG, toxicity considerations). Use numbered new lines as above.
- "guideline_primary": German S3 priority list with each item on its own line, format:
  "1) S3 [Disease], Empfehlung Nr. X.Y: \\"short quote\\""
- "guideline_secondary": secondary brief support with each item on its own line, format:
  "1) NCCN v.2025.1: short point" or "2) ESTRO/ICRU: short point"
- "key_characteristics": numbered list (each on its own line) stating:
  [Therapieelement] – Indikation: [kurze Begründung basierend auf Patientendaten/Guidelines]
  e.g., "1) RT 70 Gy – Indikation: definitive Behandlung bei lokal fortgeschrittenem HNSCC (cT3N2bM0)"
- "self_score": integer 0–100 reflecting confidence in the suggestions given the available data
  (100 = guideline-clear with complete info; lower if key data are missing/ambiguous). Return only the number.

Rules:
- Output JSON ONLY (no prose before/after).
- If critical info is missing, say what's needed in "notes_to_clinician" and give a conservative provisional plan.

Patient INPUT (verbatim):
randID: {rand_id}
Age: {age}
Diagnose: {diagnose}
Nebendiagnosen: {nebendiagnosen}
Anamnese: {anamnese}
""".strip()
\end{verbatim}

\begin{verbatim}
def build_user_content(question_text: str, image_path: str | None):
    parts = [
        {"type": "input_text", "text": (
            "Please answer the following ACR multiple-choice exam question. "
            "First answer comprehensively deriving from your expert knowledge, "
            'then give the final answer in the following form: "Final answer: X" '
            "where X is A, B, C, or D.\n\n"
            f"ACR question:\n{question_text}\n"
        )}
    ]
    if image_path:
        with open(image_path, "rb") as f:
            b64 = base64.b64encode(f.read()).decode("ascii")
        parts.append({
            "type": "input_image",
            "image_url": f"data:image/png;base64,{b64}",
        })
    return parts
\end{verbatim}
\newpage

\begin{table}[htbp]
\centering
\caption*{Supplementary Table~S1: Full list of benchmark cases included in the real-world oncologic decision-support evaluation.}
\begin{tabular}{rll}
\hline
\textbf{Case \#} & \textbf{Tumor Site} & \textbf{Clinical Scenario} \\
\hline
1  & Rectum-Anal   & Neoadjuvant -- Rectal Cancer \\
2  & Rectum-Anal   & Neoadjuvant -- Rectal Cancer \\
3  & Lung          & NSCLC -- SBRT \\
4  & Breast        & Adjuvant -- nodal positive \\
5  & Brain         & Meningioma \\
6  & Breast        & Adjuvant -- nodal positive \\
7  & Prostate      & Definitive -- low risk \\
8  & Rectum-Anal   & Neoadjuvant -- Rectal Cancer \\
9  & Lung          & SCLC \\
10 & Rectum-Anal   & Neoadjuvant -- Rectal Cancer \\
11 & Metastases    & Brain metastases \\
12 & Lung          & NSCLC -- Definitive -- Stage III \\
13 & Breast        & DCIS \\
14 & Prostate      & Definitive -- high risk \\
15 & Brain         & Glioma grade 2 / 3 \\
16 & Brain         & Vestibular Schwannoma \\
17 & Breast        & DCIS \\
18 & Brain         & Pituitary Adenoma \\
19 & Lung          & NSCLC -- Definitive -- Stage III \\
20 & Rectum-Anal   & Anal Cancer \\
21 & Breast        & Loco-regional recurrence \\
22 & Brain         & Glioblastoma grade 4 \\
23 & Rectum-Anal   & Local Recurrence \\
24 & Brain         & Vestibular Schwannoma \\
25 & Prostate      & Definitive -- high risk \\
26 & Lung          & NSCLC -- Re-RT \\
27 & Prostate      & Local recurrence after RPE + EBRT \\
28 & Rectum-Anal   & Local Recurrence \\
29 & Lung          & NSCLC -- SBRT \\
30 & Prostate      & Local recurrence after RPE + EBRT \\
31 & Brain         & Glioblastoma grade 4 \\
32 & Prostate      & Definitive -- low risk \\
33 & Lung          & NSCLC -- SBRT \\
34 & Prostate      & Definitive -- intermediate risk \\
35 & Brain         & Pituitary Adenoma \\
36 & Metastases    & SBRT \\
37 & Lung          & NSCLC -- Re-RT \\
38 & Metastases    & Brain metastases \\
39 & Metastases    & SBRT \\
40 & Lung          & NSCLC -- Definitive -- Stage III \\
41 & Prostate      & Biochemical recurrence after RPE \\
42 & Metastases    & Brain metastases \\
43 & Breast        & Adjuvant -- low-risk \\
44 & Lung          & SCLC \\
45 & Metastases    & Palliative -- bone metastases \\
46 & Prostate      & Biochemical recurrence after RPE \\
47 & Rectum-Anal   & Local Recurrence \\
48 & Metastases    & Palliative -- bone metastases \\
49 & Metastases    & Palliative -- bone metastases \\
50 & Metastases    & Brain metastases \\
51 & Rectum-Anal   & Anal Cancer \\
52 & Prostate      & Definitive -- intermediate risk \\
53 & Breast        & Adjuvant -- nodal negative \\
54 & Breast        & Adjuvant -- nodal negative \\

\hline
\end{tabular}
\end{table}

\newpage
\begin{table}[htbp]
\centering
\caption*{Supplementary Table~S1: Full list of benchmark cases included in the real-world oncologic decision-support evaluation.}
\begin{tabular}{rll}
\\
\hline
55 & Rectum-Anal   & Anal Cancer \\
56 & Breast        & Adjuvant -- low-risk \\
57 & Brain         & Glioma grade 2 / 3 \\
58 & Metastases    & SBRT \\
59 & Brain         & Meningioma \\
60 & Breast        & Loco-regional recurrence \\
\hline
\end{tabular}
\end{table}

\end{document}